\documentclass[10pt, conference, compsocconf]{IEEEtran}
\usepackage{times}
\usepackage{epsfig}
\usepackage{graphicx}
\usepackage{amsmath}
\usepackage{amssymb}
\usepackage{subfigure}
\usepackage{colortbl}
\usepackage{multirow}
\usepackage[table,xcdraw]{xcolor}
\usepackage{slashbox}

\usepackage{setspace}  

\ifCLASSINFOpdf
\else
\fi
\begin{document}
%
\title{Biometric Signature Verification \\ Using Recurrent Neural Networks}

\author{\IEEEauthorblockN{Ruben Tolosana, Ruben Vera-Rodriguez, Julian Fierrez and Javier Ortega-Garcia}
\IEEEauthorblockA{Biometrics and Data Pattern Analytics (BiDA) Lab - ATVS, Escuela Politecnica Superior, Universidad Autonoma de Madrid \\Avda. Francisco Tomas y Valiente, 11 - Campus de Cantoblanco - 28049 Madrid, Spain \\ Email: (ruben.tolosana, ruben.vera, julian.fierrez, javier.ortega)@uam.es}}


%


\maketitle

\begin{abstract}
Architectures based on Recurrent Neural Networks (RNNs) have been successfully applied to many different tasks such as speech or handwriting recognition with state-of-the-art results. The main contribution of this work is to analyse the feasibility of RNNs for on-line signature verification in real practical scenarios. We have considered a system based on Long Short-Term Memory (LSTM) with a Siamese architecture whose goal is to learn a similarity metric from pairs of signatures. For the experimental work, the BiosecurID database comprised of 400 users and 4 separated acquisition sessions are considered. Our proposed LSTM RNN system has outperformed the results of recent published works on the BiosecurID benchmark in figures ranging from 17.76\% to 28.00\% relative verification performance improvement for skilled forgeries. 
\end{abstract}


%
\IEEEpeerreviewmaketitle

\section{Introduction}
New trends based on the use of RNNs are becoming more and more important nowadays for modelling sequential data with arbitrary length \cite{DLearning_overview}. The range of applications can be very varied, from speech recognition \cite{speech_RNNs} to biomedical problems \cite{EEG_RNNs}. RNNs are defined as a connectionist model containing a self-connected hidden layer. One benefit of the recurrent connection is that a “memory” of previous inputs remains in the network’s internal state, allowing it to make use of past context. One of the fields in which RNNs has caused more impact in the last years is in handwriting recognition due to the relationship that exists between current inputs and past context. However, the range of contextual information that standard RNNs can access is very limited \cite{online_offline_handwriting} due to the well known vanishing gradient problem \cite{vanishing_gradient}. LSTM \cite{LSTM_first_article} is a RNN architecture that arised with the aim of resolving the shortcomings of standard RNNs. This architecture has been deployed with success in both on-line and off-line handwriting \cite{online_offline_handwriting, offline_handwriting}. Whereas off-line scenarios consider information only related to the image of the handwriting, in on-line scenarios additional information such as \textit{X} and \textit{Y} pen coordinates and pressure time functions are also considered providing therefore much better results. In \cite{online_offline_handwriting}, the authors proposed a system based on the use of Bidirectional LSTM (BLSTM) for recognizing unconstrained handwritten text considering both off- and on-line handwriting approaches. The results obtained applying this new approach outperformed a state-of-the-art HMM-based system and also proved the new approach to be more robust to changes in dictionary size. LSTM approaches have been considered not only for recognizing unconstrained handwriting but also for writer identification. In \cite{end_to_end_writer_identification}, the authors considered a system based on BLSTM for on-line text-independent writer identification. The experiments carried out over both English (133 writers) and Chinese (186 writers) outperformed state-of-the-art systems as well.

\begin{figure*}[htb]
\centering
\subfigure[Genuine case]{\label{figGenPen}
\includegraphics[width=0.40\linewidth]{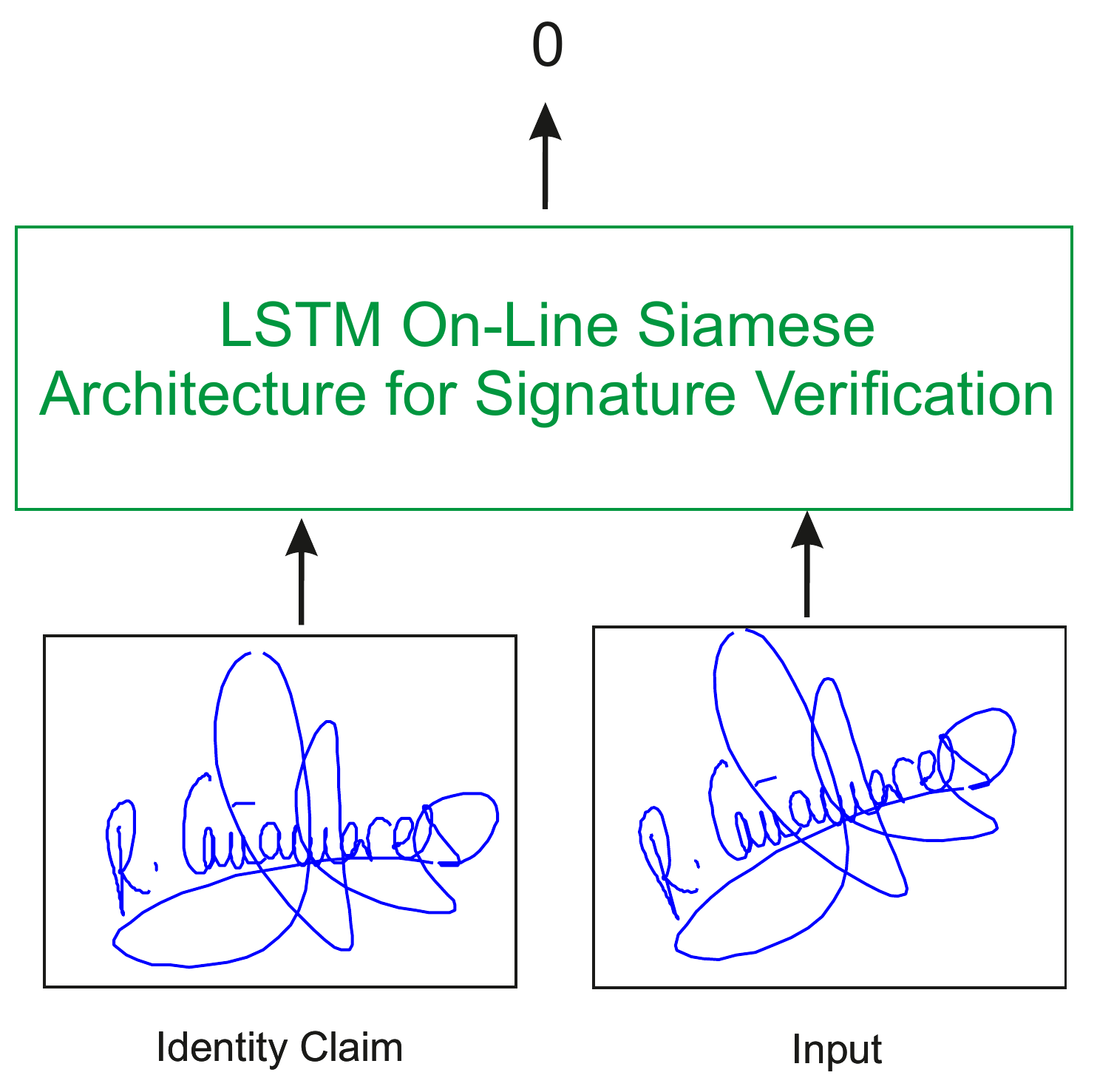}}
\hspace{0.5cm}
\subfigure[Impostor case]{\label{figGenPen}
\includegraphics[width=0.42\linewidth]{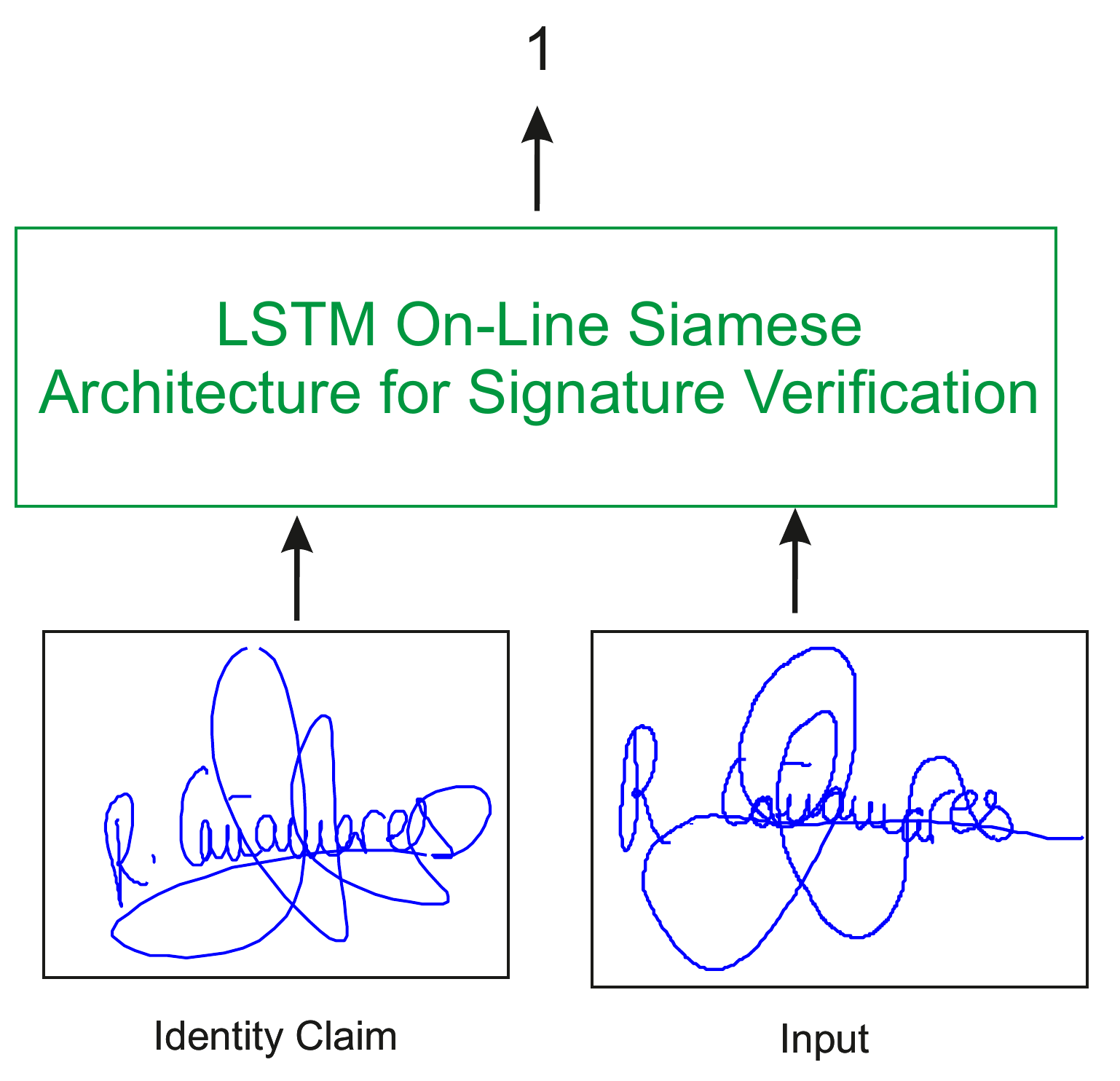}}
\caption{Examples of our proposed LSTM RNN system based on a Siamsese architecture for minimizing a discriminative cost function that drives the similarity metric to be small for pairs of signatures.} \label{fig:siamese_architecture}
\end{figure*}

Despite the good results obtained in the field of handwriting and the similarity with the case of handwritten signature, very few studies have applied LSTM RNNs successfully to handwritten signature verification systems, as far as we know. In \cite{LSTM_signature}, the authors proposed the use of a system based on LSTM for on-line signature verification. Different configurations based on the use of forget gates and peephole connections were studied considering in the experimental work a small database with only 51 users. The LSTM RNNs proposed in that work seemed to authenticate genuine and impostor cases very well. However, as it was pointed out in \cite{LSTM_signature_Liwicki}, the method proposed in that work for training the LSTM RNNs is not feasible for real applications for various reasons. First, the authors considered the same users for both development and evaluation of the system. Moreover, the deployment of that LSTM RNN architecture may not be feasible in real scenarios as the system should be trained every time a new user was enrolled in the application. In addition, forgeries are required in that approach for training, which may not be feasible to get as well. Besides, the results obtained in \cite{LSTM_signature} cannot be compared to any state-of-the-art signature verification system as the traditional measures such as the equal error rate (EER), accuracy, or calibrated log likelihood-ratios were not considered. Instead, they just reported the errors of the LSTM-outputs. In order to find some light on the feasibility of LSTM RNNs for signature verification purposes, Otte \textit{et al.} performed in \cite{LSTM_signature_Liwicki} a deep analysis considering three different real scenarios: 1) training a general network to distinguish forgeries from genuine signatures on a large training set, 2) adopting a network that works perfectly on the training set to a specific writer, and 3) training the network on genuine signatures only. However, all experiments failed obtaining a 23.75\% EER for the best configuration, far away from the best state-of-the-art results and concluding that LSTM RNN systems trained with standard mechanisms were not appropriate for the task of signature verification. 

The main contribution of this work is to analyse and prove the feasibility of LSTM RNN systems in combination with a Siamese architecture \cite{Learning_similarity_CVPR} for on-line signature verification. This Siamese architecture allows to get a close approximation to the verification task learning a similarity metric from pairs of signatures (pairs of signatures from the same user and pairs of genuine-forgery signatures). The main advantage of this method is that the model can be extrapolated to signatures from unknown users with very good results, opposite to traditional architectures where signatures from all users have to be taken into account in the training and testing process of the network in order to achieve good results \cite{LSTM_signature}. Different users and signatures are considered for the development and evaluation of the system in order to analyze the true potential of LSTM RNNs in signature verification. 

The remainder of the paper is organized as follows. In Sec. \ref{sec:proposed_methods}, our proposed approach based on the use of LSTM RNNs for signature verification is described. Sec. \ref{sec:signature_database} describes the BiosecurID on-line signature database considered in the experimental work. Sec. \ref{time_functions} describes the information used for feeding the LSTM RNNs. Sec. \ref{sec:experimental_work} describes the experimental protocol and the results achieved with our proposed approach. Finally, Sec. \ref{sec:conclusions} draws the final conclusions and points out some lines for future work.

\section{Proposed Methods}\label{sec:proposed_methods}
The methods proposed in this work for improving the performance of on-line signature verification are based on the combination of LSTM RNNs with a Siamese architecture.

\subsection{Siamese Architecture}\label{siamese_architecture}
The Siamese architecture has been used for recognition or verification applications where the number of categories is very large and not known during training, and where the number of training samples for a single category is very small \cite{Learning_similarity_CVPR}. The main goal of this architecture is to learn a similarity metric from data minimizing a discriminative cost function that drives the similarity metric to be small for pairs of signatures. Fig. \ref{fig:siamese_architecture} shows examples of the architecture proposed in this work for discriminating genuine from impostor cases. Siamese architectures have been considered for many recognition and verification applications. In \cite{Learning_similarity_CVPR}, the authors proposed the use of Convolutional Neural Networks (CNNs) with a Siamese architecture for face verification. Experiments were performed with several databases obtaining very good results where the number of training samples for a single category was very small. Siamese architectures have also been used in early works for on-line signature verification \cite{signature_verification_RNN_1993} although not considering RNNs. In \cite{signature_verification_RNN_1993}, the authors proposed an on-line signature verification system comprised of two separated sub-networks based on Time Delay Neural Networks (TDNNs). Different architectures regarding the number and size of layers were studied. A total of 8 time functions fixed to the same length of 200 points were extracted for \textit{X} and \textit{Y} pen coordinates using an old-fashion 5990 Signature Capture Device. The best performance was obtained using two convolutional layers with 12 by 64 units in the first layer and 16 by 19 units in the second one. The threshold was set to detect 80.0\% of forgeries and 95.5\% of genuine signatures, far away from the results that we can achieve nowadays with state-of-the-art systems \cite{moises_1signature, SVM_sparse}.

\begin{figure*}[tb]
  \centering
    \includegraphics[width=0.75\linewidth]{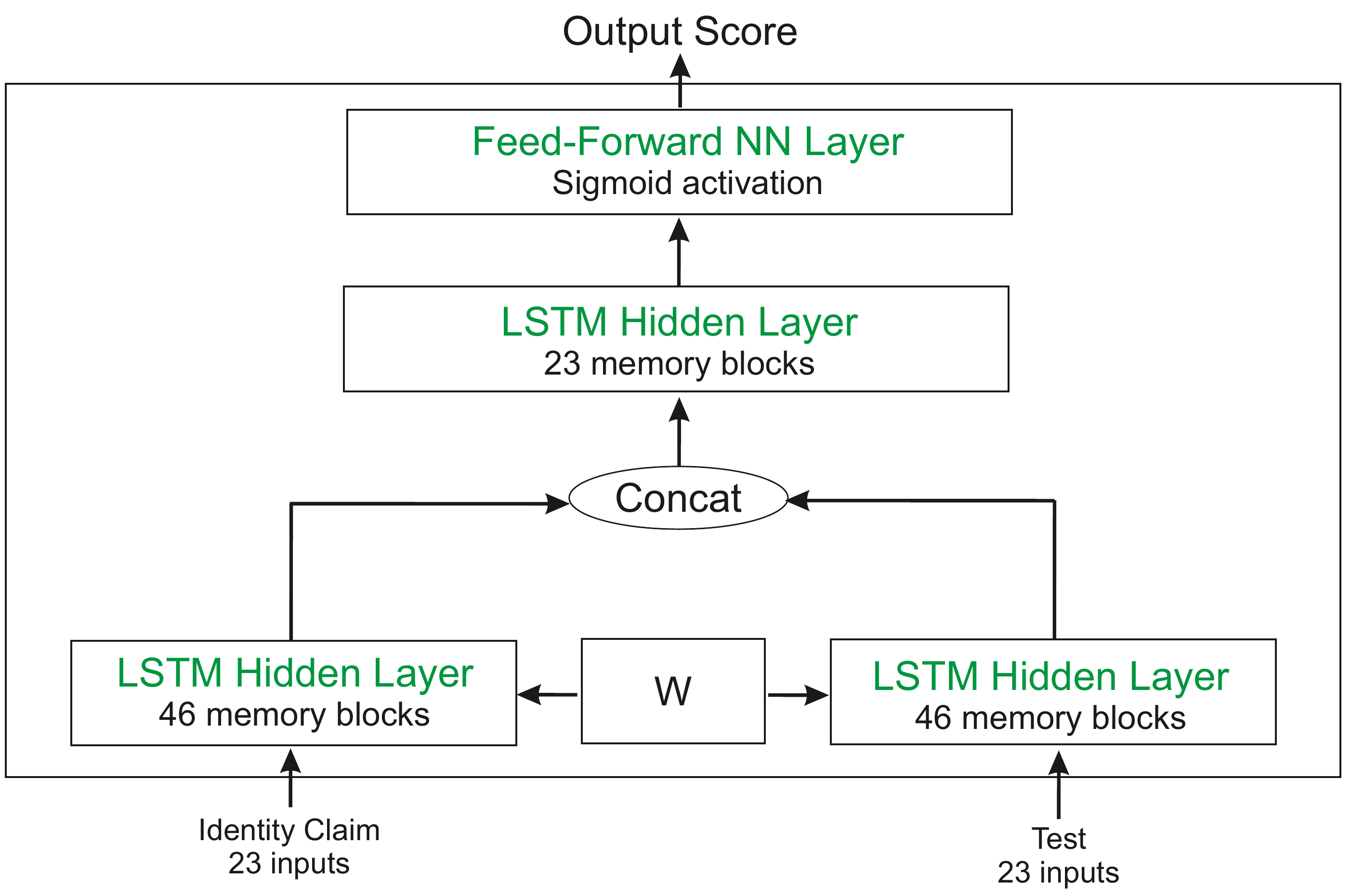}
  \caption{End-to-end on-line signature verification system proposed in this work and based on the use of LSTM RNNs with a Siamese architecture.}
  \label{fig:LSTM_configuration}
\end{figure*}

\subsection{Long Short-Term Memory}\label{LSTM_section}
LSTM RNN systems have been successfully applied to many different tasks such as language identification considering short utterances \cite{Zazo_LSTM} or biomedical problems \cite{EEG_RNNs} for example. However, the analysis and design of LSTM RNN architectures for new tasks are not straightforward \cite{how_construct_DRNN}.

LSTM RNNs \cite{LSTM_first_article} are comprised of memory blocks usually containing one memory cell each of them, a forget gate $f$, an input gate $i$, and an output gate $o$. For a time step $t$:

\begin{equation}
f_t = \sigma (W_f \cdot [h_{t-1},x_t] + b_f)
\end{equation} 

\begin{equation}
i_t = \sigma (W_i \cdot [h_{t-1},x_t] + b_i)
\end{equation} 

\begin{equation}
o_t = \sigma (W_o \cdot [h_{t-1},x_t] + b_o)
\end{equation} 

\begin{equation}
\widetilde{C_t} = \tanh (W_C \cdot [h_{t-1},x_t] + b_C)
\end{equation} 

\begin{equation}
C_t = f_t \odot C_{t-1} + i_t \odot \widetilde{C_t}
\end{equation} 

\begin{equation}
h_t = o_t \odot \tanh (C_t)
\end{equation} 

\noindent{where $W_*$ is the input-to-hidden weight matrix and $b_*$ is the bias vector. The symbol $\odot$ represents a pointwise product whereas $\sigma$ is a sigmoid layer which outputs values between 0 and 1. The LSTM does have the ability to remove old information from $t-1$ time or add new one from $t$ time. The key is the cell state $C_t$ which is carefully regulated by the gates. The $f$ gate decides the amount of previous information that passes to the new state of the cell $C_t$. The $i$ gate indicates the amount of new information (i.e. $\widetilde{C_t}$) to update in the cell state $C_t$. Finally, the output of the memory block $h_t$ is a filtered version of the cell state $C_t$, being the $o$ gate in charged of it.
}

The best topology obtained for our proposed LSTM RNNs is based on the use of two LSTM hidden layers and finally, a feed-forward neural network layer. Fig. \ref{fig:LSTM_configuration} shows our proposed end-to-end on-line signature verification system. The first layer is composed of two LSTM hidden layers with 46 memory blocks each and sharing the weights between them. The outputs provided for each LSTM hidden layer of the first layer are then concatenated and serve as input of the second layer which corresponds to a LSTM hidden layer with 23 memory blocks. Finally, a feed-forward neural network layer with a sigmoid activation is considered, providing an output score for each pairs of signatures. The size of the input layer was determined by the data, which is described in more details in Sec. \ref{time_functions}. In addition, many more LSTM RNN architectures regarding the number of LSTM hidden layers and memory blocks were tested providing worse results in all cases.

\section{On-Line Signature Database}\label{sec:signature_database}
The BiosecurID database \cite{Fierrez2009_PAA} is considered in the experimental work of this paper. This database is comprised of 16 original signatures and 12 skilled forgeries per user, captured in 4 separate acquisition sessions leaving a two-month interval between them. There are a total of 400 users and signatures were acquired considering a controlled and supervised office-like scenario. Users were asked to sign on a piece of paper, inside a grid that marked the valid signing space, using an inking pen. The paper was placed on a Wacom Intuos 3 pen tablet that captured the following time signals of each signature:  \textit{X} and \textit{Y} pen coordinates (0.25 mm), pressure (1024 levels) and timestamp (100 Hz). In addition, pen-ups trajectories are available. All the dynamic information is stored in separate text files following the format used in the first Signature Verification Competition, SVC \cite{SVC_2004}. All the acquisition process was supervised by a human operator whose task was to ensure that the collection protocol was strictly followed and that the captured samples were of sufficient quality (e.g. no part of the signature outside the designated space), otherwise, the donor was asked to repeat a given signature. 

\begin{table}[tb]
\centering
\caption{\textit{Set of time functions considered in this work.}}
\begin{tabular}{p{1cm} p{7cm}}
\hline
\# & Feature \\
\hline \hline
1 & x-coordinate: $x_n$  \\
\hline
2 & y-coordinate: $y_n$  \\
\hline
3 & Pen-pressure: $z_n$ \\
\hline
4 & Path-tangent angle: $\theta_n$  \\
\hline
5 & Path velocity magnitude: $v_n$ \\
\hline
6 & Log curvature radius: $\rho_n$ \\
\hline
7 & Total acceleration magnitude: $a_n$ \\
\hline
8-14 & First-order derivate of features 1-7: $\dot{x_n},\dot{y_n},\dot{z_n},\dot{\theta_n},\dot{v_n},\dot{\rho_n},\dot{a_n}$ \\
\hline
15-16 & Second-order derivate of features 1-2: $\ddot{x_n},\ddot{y_n}$ \\
\hline
17 & Ratio of the minimum over the maximum speed over a 5-samples window: $v^r_n$ \\
\hline
18-19 & Angle of consecutive samples and first order difference: $\alpha_n$, $\dot{\alpha_n}$ \\
\hline
20 & Sine: $s_n$ \\
\hline
21 & Cosine: $c_n$ \\
\hline
22 & Stroke length to width ratio over a 5-samples window: $r^5_n$ \\
\hline
23 & Stroke length to width ratio over a 7-samples window: $r^7_n$ \\
\hline
\end{tabular}
\label{tabla:tablaLocalFeatures}
\end{table}

\section{Time Functions Representation}\label{time_functions}
The on-line signature verification system proposed in this work is based on time functions (a.k.a. local system) \cite{eBioSign_journal}. For each signature acquired, signals related to \textit{X} and \textit{Y} pen coordinates and pressure are used to extract a set of 23 time functions, similar to \cite{2015_WIFS_SignatureHMMUpdate_RubenT} (see Table \ref{tabla:tablaLocalFeatures}). Different approaches regarding the preprocessing of the signatures and the number of time functions to consider have been analysed in a first stage. The best results are obtained feeding the LSTM RNNs with as much information as possible (i.e. 23 time functions) as it is shown in the input layer of Fig. \ref{fig:LSTM_configuration}.

\section{Experimental Work}\label{sec:experimental_work}

%
%
%
%

\subsection{Experimental Protocol}\label{sec:experimental_protocol}

\begin{figure*}[tb]
  \centering
    \includegraphics[width=1\linewidth]{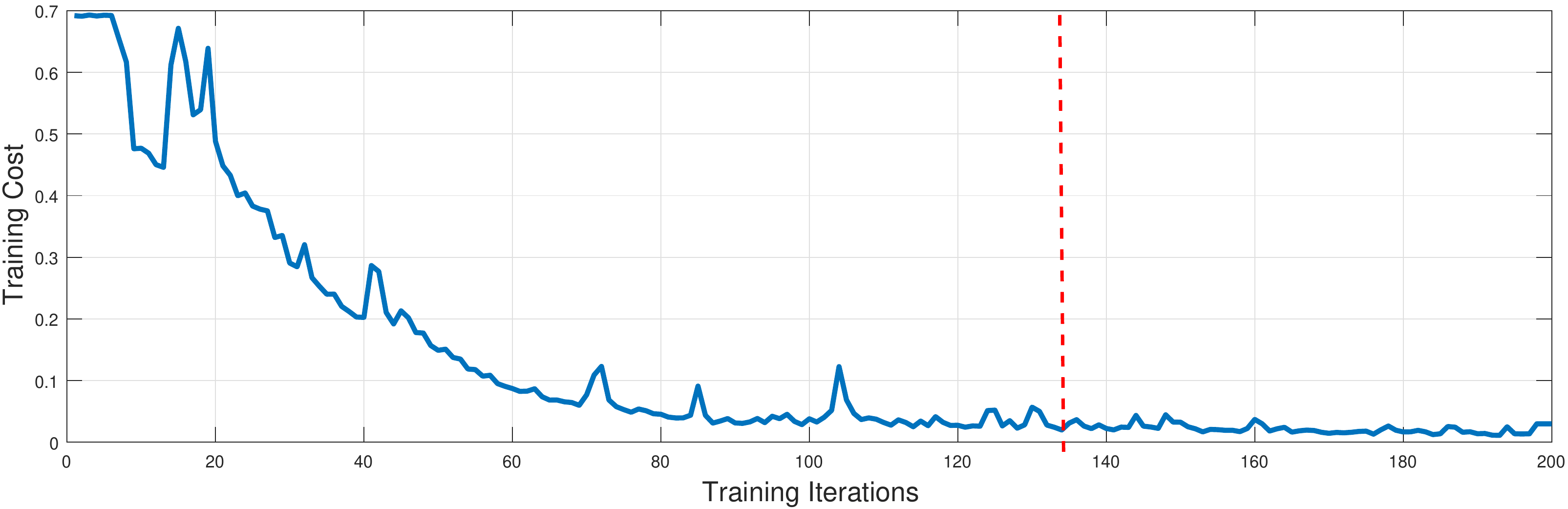}
  \caption{LSTM RNNs cost during training. The red dashed line indicates the best configuration obtained.}
  \label{fig:cost_epochs}
\end{figure*}

The experimental protocol considered in this work has been designed in order to analyse and prove the feasibility of LSTM RNNs for on-line signature verification in practical scenarios. Therefore, different users and signatures are considered for the two main stages, i.e., development of the LSTM RNNs system and evaluation of it. This allows us to obtain a clear analysis of the feasibility of these new approaches in on-line signature verification systems.

The first 300 users of the BiosecurID database are used for the development of the system, while the remaining 100 users are considered for the evaluation. For both stages, the 4 genuine signatures of the first session are used as training signatures, whereas the 12 genuine signatures of the remaining sessions are left for testing. Therefore, inter-session variability is considered in our experiments. Skilled forgeries scores are obtained by comparing training signatures against the 12 available skilled forgeries signatures for the same user.





\subsection{Experimental Results}\label{sec:experimental_results}

\subsubsection{\textbf{Development Results}}\label{sec:development}
This section describes the development and training of our proposed LSTM RNNs system with a Siamese architecture considering the 300 users of the development dataset. Two different cases are analysed: 1) the case of considering two signatures performed for the same user as inputs, and 2) the case of having one genuine signature from the claimed user and one skilled forgery signature performed by an impostor as inputs. Therefore, for the first case, a total of $4 \times 12 \times 300 = 14,400$ pairs of genuine comparisons are considered for training the system whereas for the second case, there are a total of $4 \times 12 \times 300 = 14,400$ pairs of impostor comparisons as we have the same number of genuine and skilled forgery signatures for testing. Our LSTM RNNs is implemented under Theano \cite{theano} with a NVIDIA GeForce GTX 970 GPU. Each training iteration takes about 30 minutes.

Fig. \ref{fig:cost_epochs} shows how the training cost of the LSTM RNNs decreases with the number of training iterations. A red dashed line is included in the figure indicating the training iteration which provides the best LSTM RNN performance over the development dataset, with a training cost value of 0.019. It is important to remark the behaviour of the neural network during training as it is capable of skipping different local minimums during the training process and continue decreasing the training cost until about 140 training iterations where it saturates. Regarding the system performance, two different cases are considered. First, the evaluation of the system performance considering scores directly from all pairs of signatures (i.e. 1vs1) and second, the case of performing the average score of the four one-to-one comparisons (i.e. 4vs1) as there are four genuine training signatures per user. Our proposed LSTM RNN system achieves a system performance in training of 0.11\% and 0.00\% EER for the cases 1vs1 and 4vs1 respectively. These results shows the potential of LSTM RNNs for signature verification.

\subsubsection{\textbf{Evaluation Results}}\label{sec:evaluation}
This section analyses the performance of the proposed LSTM RNNs trained in the previous section. The remaining 100 users (not used for development) are considered here. In order to make comparable our approach to related works, we have used the same Baseline System recently considered in \cite{2015_ICB_skilledSignSigmaLog_Marta}, which is based on the DTW algorithm with a total of 9 out of 27 different time functions selected using the Sequential Forward Feature Selection (SFFS) algorithm.

\begin{figure}[tb]
  \centering
    \includegraphics[width=0.85\linewidth]{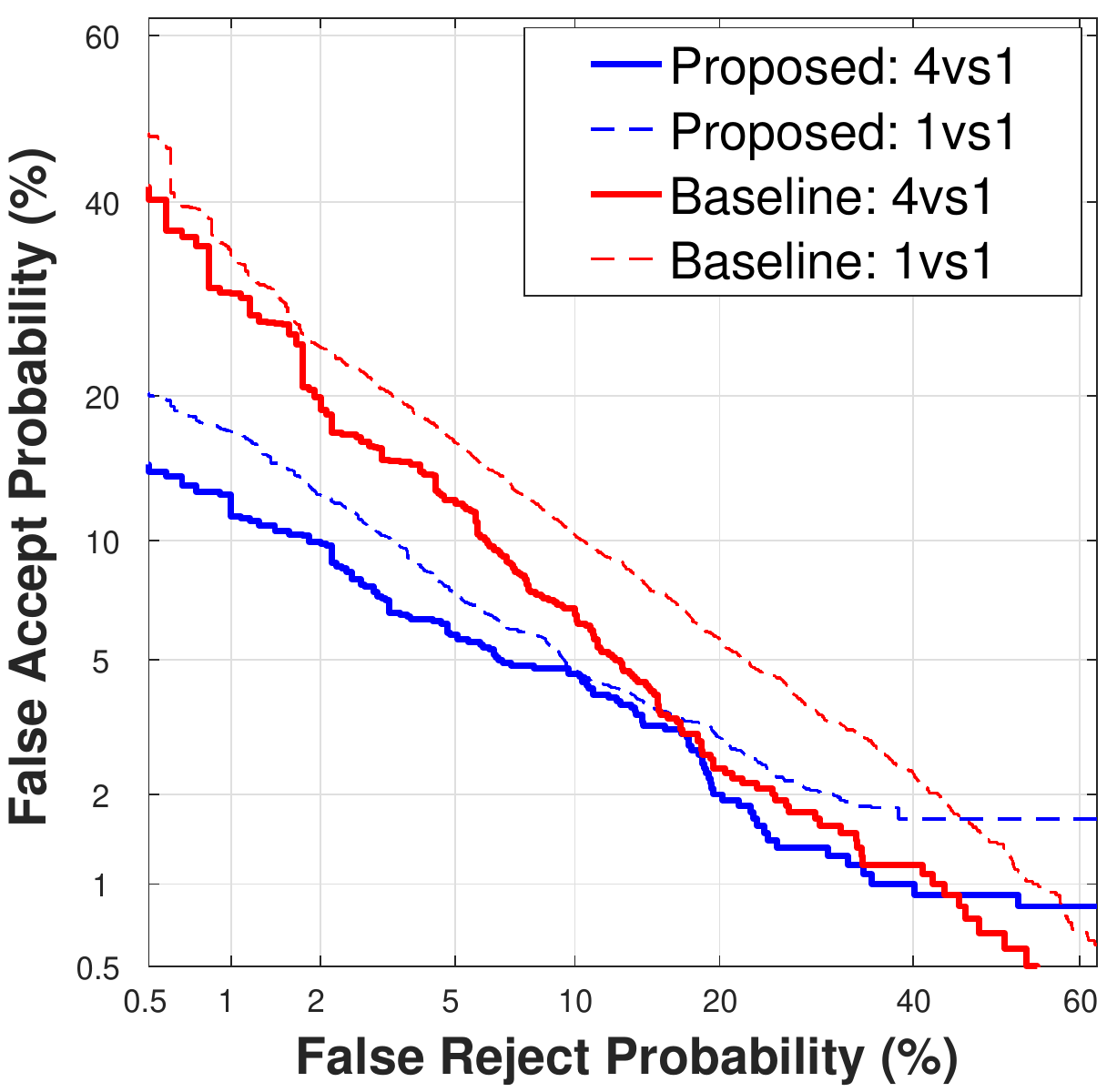}
  \caption{System performance results for both Proposed and Baseline Systems and both 1vs1 and 4vs1 cases considering the evaluation dataset.}
  \label{fig:evaluation}
\end{figure}

Fig. \ref{fig:evaluation} shows the system performance in terms of DET curves for both Proposed and Baseline Systems considering 1vs1 and 4vs1 cases. Table \ref{table_eer_evaluation} shows the system performance in terms of EER(\%) for completeness.

Analysing the results obtained in Table \ref{table_eer_evaluation} for the 1vs1 case, our Proposed System achieves a relative improvement of 36.7\% EER compared to the Baseline System. This result (i.e. 6.44\% EER) outperforms state-of-the-art results for the case of considering one signature for training \cite{moises_1signature}. 

\begin{table}[]
\centering
\caption{System performance results in terms of EER(\%) \break considering the evaluation dataset.}
\label{table_eer_evaluation}
\begin{tabular}{c|c|c|}
\cline{2-3}
                           & 1vs1 & 4vs1 \\ \hline
\multicolumn{1}{|c|}{Baseline System} & 10.17           & 7.75 \\ \hline
\multicolumn{1}{|c|}{Proposed System} & 6.44                      & 5.58            \\ \hline
\end{tabular}
\end{table}

Analysing the results obtained for the 4vs1 case, our Proposed System achieves a relative improvement of 28.0\% EER compared to the Baseline System, outperforming the state-of-the-art results of the BiosecurID database with a final value of 5.58\% EER. Moreover, it is important to highlight that the result obtained with our Proposed System for the case of using just one training signature (1vs1) outperforms the result obtained with the Baseline System for the 4vs1 case, showing the high ability of our proposed approach for learning even with small amounts of data. 

Results obtained prove the high feasibility of our proposed LSTM RNNs with a Siamese architecture for on-line signature verification. In addition, it is important to highlight the advantages of considering our proposed approach for the deployment in real applications as the LSTM RNN system does not require any kind of training during the evaluation stage and works independently of the number of training signatures available for the user.

Finally, a preliminary evaluation considering random forgeries has been carried out for completeness. It is important to highlight that our proposed LSTM RNN system has been developed only for skilled and not for random forgeries as this is the most challenging case in real scenarios not only because of the quality of the forgeries but to the scarce of skilled forgeries. The same experimental protocol considered for the evaluation of skilled forgeries (i.e. the remaining 100 users) is carried out for the case of random forgeries, but comparing the reference signatures with one genuine signature of each of the remaining users. The system performance obtained for this case has been around 24.0\% EER, much higher compared to the 0.5\% EER obtained using the Baseline System based on DTW. This result achieved using our proposed LSTM RNN system makes sense as the Siamese architecture has learnt the similarity metric from pairs of signatures minimizing a discriminative cost function only for the skilled forgeries case. Therefore, the same very good results are also expected to be achieved for the case of random forgeries when pairs of genuine and random forgeries are considered during the development and training of the system. In addition, it is always feasible to perform an on-line signature verification system based on two consecutive stages: 1) a system based on DTW in order to reject random forgeries, and 2) a system based on our proposed LSTM RNNs system in order to reject skilled forgeries. This way we would achieve state-of-the-art results for both skilled and random forgery cases.

\section{Conclusions}\label{sec:conclusions}
In this work we analyse and prove the feasibility of LSTM RNN systems in combination with a Siamese architecture \cite{Learning_similarity_CVPR} for on-line signature verification. This work provides the first successful framework on the use of RNN systems for on-line signature verification, as far as we know. The BiosecurID database comprised of 400 users and 4 separated acquisition sessions has been considered in the experimental work, using the first 300 users for development and the remaining 100 users for evaluation.  Two different cases have been considered. First, the evaluation of the system performance considering scores directly from all pairs of signatures (i.e. 1vs1) and second, the case of performing the average score of the four one-to-one comparisons (i.e. 4vs1) as there are 4 genuine training signatures per user (from the first session).

Our proposed LSTM RNN system with a Siamese architecture is based on two LSTM hidden layers and finally a feed-forward neural network with a sigmoid activation. The best model has obtained in development a final value of 0.11\% and 0.0\% EER for the 1vs1 and 4vs1 cases, respectively. 

Analysing the results obtained using the 100 users of the evaluation dataset, our Proposed System has achieved a final value of 6.44\% and 5.58\% EER for the 1vs1 and 4vs1 cases respectively. These results have outperformed the state-of-the-art either for the case of using just one training signature (1vs1) \cite{moises_1signature} or the case of performing the average score of the four one-to-one comparisons (4vs1) \cite{2015_ICB_skilledSignSigmaLog_Marta}. In addition, it is important to highlight the results obtained in this work compared to the ones obtained by Otte \textit{et al.} in \cite{LSTM_signature_Liwicki} where all experiments failed obtaining a 23.75\% EER for the best case. In that work, standard LSTM architectures seemed not to be appropriate for the task of signature verification. However, our proposed Siamese architecture allows to get a close approximation to the verification task learning a similarity metric from pairs of signatures (pairs of signatures from the same user and pairs of genuine-forgery signatures).

These results prove the high feasibility of our proposed LSTM RNNs with a Siamese architecture for on-line signature verification. For future work, the approach considered in this work will be further analysed considering not only skilled but random forgeries during the training of the neural network. 

\section*{Acknowledgments}
This work has been supported by project TEC2015-70627-R MINECO/FEDER and by UAM-CecaBank Project. Ruben Tolosana is supported by a FPU Fellowship from Spanish MECD.

\bibliographystyle{IEEEtran}
\bibliography{egbib2}

\end{document}